\documentclass[11pt, a4paper]{article}
\usepackage{hyperref}
\usepackage[hyperref]{acl2019}
\usepackage{url}
\usepackage{times}
\usepackage{latexsym}
\usepackage{wrapfig}

\usepackage{graphicx}
\usepackage{subcaption}
\usepackage{amsmath,amssymb}
\usepackage{physics}
\usepackage{epstopdf}
\usepackage{xspace}
\usepackage{xcolor}
\usepackage{verbatim}
\usepackage{tikz}
\usetikzlibrary{bayesnet}
\usepackage{scrextend}
\usepackage{booktabs}
\usepackage{tabularx}
\usepackage{rotating}
\usepackage{capt-of}
\usepackage{multicol}

\definecolor{darkblue}{rgb}{0, 0, 0.5}
\hypersetup{colorlinks=true,citecolor=darkblue, linkcolor=darkblue, urlcolor=darkblue}

\DeclareMathOperator*{\argmax}{arg\,max}
\DeclareMathOperator*{\Cat}{Cat}

\DeclareMathOperator*{\KL}{KL}
\DeclareMathOperator*{\JS}{JS}
\DeclareMathOperator*{\D}{D}
\DeclareMathOperator*{\ELBO}{ELBO}
\DeclareMathOperator*{\diag}{diag}
\DeclareMathOperator*{\greedy}{greedy}

\DeclareMathOperator*{\softmax}{softmax}
\DeclareMathOperator*{\ReLU}{ReLU}
\DeclareMathOperator*{\softplus}{softplus}
\DeclareMathOperator*{\affine}{affine}
\DeclareMathOperator*{\BiGRU}{BiGRU}
\DeclareMathOperator*{\GRU}{GRU}
\DeclareMathOperator*{\detach}{detach}
\DeclareMathOperator*{\block}{block}
\DeclareMathOperator*{\attention}{attention}
\DeclareMathOperator*{\emb}{emb}
\DeclareMathOperator*{\avg}{avg}
\DeclareMathOperator*{\concat}{concat}

\usepackage{soul}
\usepackage{ifthen}

\usepackage{xparse}

\NewDocumentCommand{\delete}{ O{} O{} m }{%
    \ifthenelse{ \equal{#1}{} }
      {\textcolor{red}{\st{#3}}}
      {\textcolor{red}{\st{#3}$^{\text{#1}}$}}
    \ifthenelse{ \equal{#2}{} }
      {}
      {\footnote{#2}}
}
\NewDocumentCommand{\add}{ O{} O{} m }{%
    \ifthenelse{ \equal{#1}{} }
      {\textcolor{blue}{#3}}
      {\textcolor{blue}{#3$^{\text{#1}}$}}
    \ifthenelse{ \equal{#2}{} }
      {}
      {\footnote{#2}}
}
\NewDocumentCommand{\replace}{ O{} O{} m m }{%
    \delete{#3}\add[#1][#2]{#4}
}
\NewDocumentCommand{\opinion}{ O{} m m }{%
    \textcolor{orange}{#2}\footnote{\textcolor{orange}{#1: }#3}
}

\title{Auto-Encoding Variational Neural Machine Translation}

\author{Bryan Eikema \& Wilker Aziz\\
Institute for Logic, Language and Computation\\
University of Amsterdam\\
\texttt{b.eikema@uva.nl, w.aziz@uva.nl} \\ 
}
\date{}

\def\aclpaperid{349}
\aclfinalcopy

\newcommand{\y}{\ensuremath{y}}
\newcommand{\n}{\ensuremath{|y|}}
\newcommand{\yh}{\ensuremath{y_{<j}}}

\newcommand{\x}{\ensuremath{x}}
\newcommand{\m}{\ensuremath{|x|}}
\newcommand{\xh}{\ensuremath{x_{<i}}}

\begin{document}
\maketitle

\begin{abstract}
We present a deep generative model of bilingual sentence pairs for machine translation. The model generates source and target sentences jointly from a shared latent representation and is parameterised by neural networks. We perform efficient training using amortised variational inference and reparameterised gradients. Additionally, we discuss the statistical implications of joint modelling and propose an efficient approximation to maximum a posteriori decoding for fast test-time predictions. We demonstrate the effectiveness of our model in three machine translation scenarios: in-domain training, mixed-domain training, and learning from a mix of gold-standard and synthetic data. Our experiments show consistently that our joint formulation outperforms conditional modelling (i.e. standard neural machine translation) in all such scenarios. %
\end{abstract}

\section{Introduction}
\label{sec:introduction}

Neural machine translation (NMT) systems \citep{Kalchbrenner+2013:RCTM,Sutskever+2014:SSNN,Cho+2014:RNNMT} require vast amounts of  labelled data, i.e. bilingual sentence pairs, to be trained effectively. 
Oftentimes, the data we use to train these systems are a byproduct of mixing different sources of data. 
For example, labelled data are sometimes obtained by putting together corpora from different domains \citep{sennrich-EtAl:2017:WMT}. 
Even for a single domain, parallel data often result from the combination of documents independently translated from different languages by different people or agencies, possibly following different guidelines. 
When resources are scarce, it is not uncommon to mix in some synthetic data, e.g. bilingual data artificially obtained by having a model translate target monolingual data to the source language \citep{SennrichEtAl2016Mono}.
Translation direction, original language, and quality of translation are some of the many factors that we typically choose not to control for (due to lack of information or simply for convenience).\footnote{Also note that this list is by no means exhaustive. For example,  \citet{rabinovich-EtAl:2017:EACLlong} show influence of factors such as personal traits and demographics in translation.  Another clear case is presented by \citet{johnson2016google}, who combine parallel resources for multiple languages to train a single encoder-decoder architecture.}
All those arguably contribute to making our labelled data a mixture of samples from various data distributions.

Regular NMT systems do not explicitly account for latent factors of variation, 
instead, %
given a source sentence, NMT models a single conditional distribution over target sentences as a fully supervised problem.
In this work, 
we introduce a deep generative model %
that generates source and target sentences jointly from a shared latent representation. 
The model has the potential to use the latent representation to capture global aspects of the observations, such as some of the latent factors of variation just discussed.
The result is a model that accommodates members of a more complex class of marginal distributions. 
Due to the presence of latent variables, this model requires posterior inference, in particular, we employ the framework of amortised variational inference \citep{Kingma+2014:VAE}. 
Additionally, we propose an efficient approximation to maximum a posteriori (MAP) decoding for fast test-time predictions.

\paragraph{Contributions} We introduce a deep generative model for NMT (\S\ref{sec:model}) and discuss  theoretical advantages of joint modelling over conditional modelling (\S\ref{sec:stat-remarks}). 
We also derive an efficient approximation to MAP decoding that requires only a single forward pass through the network for prediction (\S\ref{sec:efficient-MAP}).
Finally, we show in \S\ref{sec:exp} that our proposed model improves translation performance in at least three practical scenarios: i) in-domain training on little data, where test data are expected to follow the training data distribution closely; ii) mixed-domain training, where we train a single model but test independently on each domain; and iii) learning from large noisy synthetic data.

\section{\label{sec:nmt}Neural Machine Translation}

In machine translation our observations are pairs of random sequences, a source sentence $x = \langle x_1, \ldots, x_m \rangle$ and a target sentence $y = \langle y_1, \ldots, y_n \rangle$, whose lengths $m$ and $n$ we denote by $|x|$ and $|y|$, respectively. 
In NMT, the likelihood of the target given the source 
\begin{equation}
\begin{aligned}
    P(\y|\x, \theta) &= \prod_{j=1}^{\n} \Cat(y_j|f_\theta(\x, \yh))
\end{aligned}
\end{equation}
factorises without Markov assumptions \citep{Sutskever+2014:SSNN,BahdanauEtAl2015NMT,ChoEtyAl14:SST}.
We have a fixed parameterised function $f_\theta$, i.e. a neural network architecture, compute categorical parameters for varying inputs, namely, the source sentence and target prefix (denoted \yh). 

Given a dataset $\mathcal D$ of i.i.d. observations, the parameters $\theta$ of the model %
are point-estimated to attain a local maximum of the log-likelihood function, $\mathcal L(\theta|\mathcal D) = \sum_{(\x, \y) \in \mathcal D} \log P(y|x, \theta)$,
via stochastic gradient-based optimisation \citep{robbinsmonro:1951,Bottou2003}.

\paragraph{Predictions} For a trained model, predictions are performed by searching for the target sentence $\y$ that maximises the conditional $P(\y|\x)$, or equivalently its logarithm, with a greedy algorithm 
\begin{equation}
    \argmax_{\y} ~ P(\y|\x, \theta) \approx \greedy_{\y} ~ \log P(\y|\x, \theta)
\end{equation}
such as beam-search \citep{Sutskever+2014:SSNN}, possibly aided by a manually tuned length penalty. %
This \emph{decision rule} is often referred to as MAP decoding \citep{Smith:2011:strucpred}.

\section{Auto-Encoding Variational NMT}
\label{sec:model}

To account for a latent space where global features of observations can be captured, we introduce a random sentence embedding $z \in \mathbb R^d$ and model the joint distribution over observations as a marginal of 
$p(z, \x, \y|\theta)$.\footnote{We use uppercase $P(\cdot)$ for probability mass functions and lowercase $p(\cdot)$ for probability density functions.} 
That is, $(\x, \y) \in \mathcal D$ is assumed to be sampled from the distribution
\begin{equation}\label{eq:joint-marginal}
    \begin{aligned}
        P(\x, \y|\theta) &= \int p(z) P(\x, \y|z, \theta)\dd z ~ .
    \end{aligned}
\end{equation}
where we impose a standard Gaussian prior on the latent variable, i.e. $Z \sim \mathcal N(0, I)$, and assume $X \perp Y | Z$. %
That is, given a sentence embedding $z$, 
we first generate the source conditioned on $z$, 
\begin{equation}
\begin{aligned}
P(\x|z, \theta) %
=&\prod_{i=1}^{\m} \Cat(x_i|g_\theta(z, \xh)) ~,
\end{aligned}
\end{equation}
then generate the target conditioned on $\x$ and $z$, 
\begin{equation}
\begin{aligned}
P(\y|\x, z, \theta) %
=& \prod_{j=1}^{\n} \Cat(y_j|f_\theta(z, \x, \yh)) ~.
\end{aligned}
\end{equation}
Note that the source sentence is generated without Markov assumptions by drawing one word at a time from a categorical distribution parameterised by a recurrent neural network $g_\theta$. The target sentence is generated similarly by drawing target words in context from a categorical distribution parameterised by a sequence-to-sequence architecture $f_\theta$. 
This essentially combines a neural language model \citep{mikolov2010recurrent} and a neural translation model (\S\ref{sec:nmt}), each extended to condition on an additional stochastic input, namely, $z$.

\subsection{\label{sec:stat-remarks}Statistical considerations} 

Modelling the conditional directly, as in standard NMT, corresponds to the statistical assumption that the \emph{distribution} over source sentences can provide no information about the distribution over target sentences given a source. 
That is, conditional NMT assumes independence of $\beta$ determining $P(y|x, \beta)$ and $\alpha$ determining $P(x|\alpha)$. 
Scenarios where this assumption is unlikely to hold are common: where $\x$ is noisy (e.g. synthetic or crowdsourced), poor quality $\x$ should be assigned low probability $P(\x|\alpha)$ which in turn should inform the conditional. Implications of this assumption extend to parameter estimation:  updates to the conditional are not sensitive to how exotic $\x$ is.  

Let us be more explicit about how we parameterise our model by identifying $3$ sets of parameters $\theta = \{\theta_{\text{emb-x}}, \theta_{\text{LM}}, \theta_{\text{TM}}\}$, where $\theta_{\text{emb-x}}$ parameterises an embedding layer for the source language. 
The embedding layer is shared  between the two model components
\begin{equation}
\begin{aligned}
P(x, y|z, \theta) = \\
P(x|\underbrace{z, \theta_{\text{emb-x}}, \theta_{\text{LM}}}_{\alpha}) P(y|x, \underbrace{z, \theta_{\text{emb-x}}, \theta_{\text{TM}}}_{\beta})    
\end{aligned}
\end{equation}
and it is then clear by inspection that $\alpha \cap \beta = \{z, \theta_{\text{emb-x}}\}$. In words, 
we break the independence assumption in two ways, namely, by having the two distributions share parameters and by having them depend on a shared latent sentence representation $z$.
Note that while the embedding layer is deterministic and global to all sentence pairs in the training data, the latent representation is stochastic and local to each sentence pair.

Now let us turn to considerations about latent variable modelling. 
Consider a model $P(\x|\theta_{\text{emb-x}}, \theta_{\text{LM}})P(\y|\x, \theta_{\text{emb-x}}, \theta_{\text{TM}})$ of the joint distribution over observations that does not employ latent variables.
This alternative, which we discuss further in experiments, models each component directly, whereas our proposed model~(\ref{eq:joint-marginal}) requires marginalisation of latent embeddings $z$. 
Marginalisation turns our directed graphical model into an undirected one inducing further structure in the marginal. %
See Appendix \ref{app:stats}, and Figure \ref{fig:AEVNMT-zoom} in particular, for an extended discussion.

\subsection{\label{sec:estimation}Parameter estimation} 
The marginal in Equation (\ref{eq:joint-marginal}) is clearly intractable, thus precluding maximum likelihood estimation.  %
Instead, we resort to variational inference \citep{Jordan+1999:VI,Blei+2016:VI}  and introduce a variational approximation $q(z|\x, \y, \lambda)$
to the intractable posterior $p(z|\x, \y, \theta)$. %
We let the approximate posterior be a diagonal Gaussian 
\begin{equation}
\begin{aligned}
    Z|\lambda, \x, \y &\sim \mathcal N(\mathbf u, \diag(\mathbf s \odot \mathbf s)) \\
    \mathbf u &= \mu_\lambda(\x, \y) \\
    \mathbf s &= \sigma_\lambda(\x, \y) \\
\end{aligned}
\end{equation}
and predict its parameters (i.e. $\mathbf u \in \mathbb R^d, \mathbf s \in \mathbb R^d_{>0}$) with neural networks whose parameters we denote by $\lambda$. This makes the model an instance of a variational auto-encoder \citep{Kingma+2014:VAE}. See Figure \ref{fig:AEVNMT} in Appendix \ref{app:stats} for a graphical depiction of the generative and inference models.

We can then jointly estimate the parameters of both models (generative  $\theta$ and inference $\lambda$) by maximising the ELBO \citep{Jordan+1999:VI}, a lowerbound on the marginal log-likelihood, 
\begin{equation}\label{eq:ELBO}
\begin{aligned}
    \log P(\x, \y|\theta) \ge \mathcal E(\theta, \lambda|\x, \y) =\\
    \mathbb E_{\epsilon \sim \mathcal N(0, I)}\left[ \log P(\x, \y|z=\mathbf u + \epsilon \odot \mathbf s, \theta) \right] \\ 
    -\KL(\mathcal N(z|\mathbf u, \diag(\mathbf s \odot \mathbf s)) || \mathcal N(z|0, I)) ~,
\end{aligned}
\end{equation}
where we have expressed the expectation with respect to a fixed distribution---a reparameterisation available to location-scale families such as the Gaussian \citep{Kingma+2014:VAE,RezendeEtAl14VAE}.  
Due to this reparameterisation, we can compute a Monte Carlo estimate of the gradient of the first term via back-propagation \citep{rumelhartetal:1986,schulmanetal:2015}.
The $\KL$ term, on the other hand, is available in closed form~\citep[Appendix B]{Kingma+2014:VAE}.

\subsection{\label{sec:efficient-MAP}Predictions} 

In a latent variable model, MAP decoding (\ref{eq:argmax-marginal-conditional}) requires searching for $\y$ that maximises the marginal  $P(\y|\x, \theta) \propto P(\x, \y|\theta)$, or equivalently its logarithm. %
In addition to approximating exact search with a greedy algorithm, other approximations are necessary in order to achieve fast prediction. 
First, rather than searching through the true marginal, we search through the evidence lowerbound. %
Second, we  replace the approximate posterior $q(z|\x, \y)$ by an auxiliary distribution $r(z|\x)$. 
As we are searching through the space of target sentences, not conditioning on $\y$ circumvents combinatorial explosion and allows us to drop terms that depend on $\x$ alone (\ref{eq:argmax-r-drop}).
Finally, instead of approximating the expectation via MC sampling, we condition on the expected latent representation and search greedily (\ref{eq:argmax-conditional}).
\begin{subequations}
\begin{align}
    &\argmax_{\y} ~ \log P(\y|\x) \label{eq:argmax-marginal-conditional} \\
    &\approx \argmax_{\y} ~ \mathbb E_{r(z|\x)}[\log P(\y|z, \x)]  \label{eq:argmax-r-drop} \\
    &\approx \greedy_{\y} ~ \log P(\y|\mathbb E_{r(z|\x)}[z], \x) \label{eq:argmax-conditional} 
\end{align}
\end{subequations}
Together, these approximations enable prediction with a single call to an $\argmax$ solver, in our case a standard greedy search algorithm, which leads to prediction times that are very close to that of the conditional model. 
This strategy, and (\ref{eq:argmax-r-drop}) in particular,  suggests that a good auxiliary distribution $r(z|\x)$ should approximate $q(z|\x, \y)$ closely.

We parameterise this \emph{prediction model} using a neural network and investigate different options to estimate its parameters.
As a first option, we restrict the approximate posterior to conditioning on $\x$ alone, i.e. we approach posterior inference with $q_\lambda(z|\x)$ rather than $q_\lambda(z|\x, \y)$, and thus,  we can use $r(z|\x) = q_\lambda(z|\x)$ for prediction.\footnote{Note that this does not stand in contrast to our motivation for joint modelling, as we still tie source and target through $z$ in the generative model, but it does limit the context available for posterior inference.} 
As a second option, we make $r_\phi(z|\x)$ a diagonal Gaussian %
and estimate parameters $\phi$ to make $r_\phi(z|\x)$  close to the approximate posterior $q_\lambda(z|\x, \y)$ as measured by $\D(r_\phi, q_\lambda)$.
For as long as $\D(r_\phi, q_\lambda) \in \mathbb R_{\ge 0}$ for every choice of $\phi$ and $\lambda$, we can estimate $\phi$ jointly with $\theta$ and $\lambda$ by maximising a modified $\ELBO$
\begin{equation}\label{eq:ELBO-D}
    \log P(\x, \y|\theta) \ge \mathcal E(\theta, \lambda|\x, \y) - \D(r_\phi, q_\lambda)
\end{equation}
which is loosened by the gap between $r_\phi$ and $q_\lambda$.
In experiments we investigate a few options for $\D(r_\phi, q_\lambda)$, all available in closed form for Gaussians, such as $\KL(r_\phi || q_\lambda)$, $\KL(q_\lambda||r_\phi)$, as well as  
the Jensen-Shannon ($\JS$) divergence. %

Note that $r_\phi$ is used only for prediction as a decoding \emph{heuristic} and as such need not be stochastic. We can, for example, design $r_\phi(\x)$ to be a point estimate of the posterior mean and optimise
\begin{equation}\label{eq:ELBO-L2}
\mathcal E(\theta, \lambda|\x, \y) - \norm{r_\phi(\x) - \mathbb E_{q_\lambda(z|\x, \y)}[z]}_2^2
\end{equation}
which remains a lowerbound on log-likelihood.
%
%

%
%
%

\section{\label{sec:exp}Experiments}

We investigate two translation tasks, namely, WMT's  translation of news \citep{WMT16} and IWSLT's translation of transcripts of TED talks \citep{cettolo2014report}, 
and concentrate on translations for German (\textsc{De}) and English (\textsc{En}) in either direction.
In this section we aim to investigate scenarios where we expect observations to be representative of various data distributions. 
As a sanity check, we start where training conditions can be considered in-domain with respect to test conditions. Though note that this does not preclude the potential for appreciable variability in observations as various other latent factors still likely play a role (see \S\ref{sec:introduction}). 
We then mix datasets from these two remarkably different translation tasks and investigate whether  performance can be improved across tasks with a single model. 
Finally, we investigate the case where we learn from synthetic data in addition to gold-standard data. 
For this investigation we derive synthetic data from observations that are close to the domain of the test set in an attempt to avoid further confounders.

\paragraph{Data} 
For bilingual data we use News Commentary (NC) v12 \citep{WMT17} and IWSLT 2014 \citep{cettolo2014report}, where we assume NC to be representative of the test domain of the WMT News task. 
The datasets consist of $255,591$ training sentences and $153,326$ training sentences respectively. 
In experiments with synthetic data, we subsample $10^6$  sentences from the News Crawl 2016 articles \citep{WMT17} for either German or English depending on the target language. 
For the WMT task, we concatenate \texttt{newstest2014} and \texttt{newstest2015} for validation/development ($5,172$ sentence pairs) and report test results on \texttt{newstest2016} ($2,999$ sentence pairs).
For IWSLT, we use the split proposed by \citet{ranzato2015sequence} who separated $6,969$ training instances for validation/development and reported test results on a concatenation of \texttt{dev2010}, \texttt{dev2012} and \texttt{tst2010-2012} ($6,750$ sentence pairs).

\paragraph{Pre-processing} We tokenized and truecased all data using standard scripts from the Moses toolkit~\citep{Koehn+2007:moses}, and removed sentences longer than $50$ tokens. For computational efficiency and to avoid problems with closed vocabularies, we segment the data using BPE~\citep{sennrich-haddow-birch:2016:P16-12} with $32,000$ merge operations independently for each language. 
For training the truecaser and the BPEs we used a concatenation of all the available bilingual and monolingual data for German and all bilingual data for English.

\paragraph{Systems} We develop all of our models on top of Tensorflow NMT \citep{luong17}.
Our baseline system is a standard implementation of conditional NMT (\textsc{Cond}) \citep{BahdanauEtAl2015NMT}.
To illustrate the importance of latent variable modelling, we also include in the comparison a simpler attempt at \textsc{Joint} modelling where we do not induce a shared latent space. Instead, the model is trained in a fully-supervised manner to maximise what is essentially a combination of two nearly independent objectives, 
\begin{multline}
    \mathcal L(\theta|\mathcal D) = \!\!\!\sum_{(\x, \y) \in \mathcal D} \sum_{i=1}^{\m} \log P(x_i|\xh, \theta_{\text{emb-x}}, \theta_{\text{LM}}) \\+ \sum_{j=1}^{\n} \log P(y_j|\x, \yh, \theta_{\text{emb-x}}, \theta_{\text{TM}}) ~,
\end{multline}
namely, a language model and a conditional translation model. 
Note that the two components of the model share very little, i.e. an embedding layer for the source language.
Finally, we aim at investigating the effectiveness of our auto-encoding variational NMT (\textsc{AEVNMT}).\footnote{Code available from \url{github.com/Roxot/AEVNMT}.} 
Appendix \ref{app:arch} contains a detailed description of the architectures that parameterise our systems.\footnote{In comparison to \textsc{Cond}, \textsc{AEVNMT} requires additional components: a source language model, an inference network, and possibly a prediction network. However, this does not add much sequential computation: the inference network can run in parallel with the source encoder, and the source language model runs in parallel with the target decoder.}

\paragraph{Hyperparameters} Our recurrent cells are $256$-dimensional GRU units~\citep{Cho+2014:RNNMT}.  
We train on batches of $64$ sentence pairs with Adam \citep{kingma2015adam}, learning rate $3\times 10^{-4}$, 
for at least $T$ updates. We then perform convergence checks every $500$ batches and stop after $20$ checks without any improvement measured by BLEU \citep{BLEU}. 
For in-domain training we set $T=140,000$, and for mixed-domain training, as well as training with synthetic data, we set $T=280,000$. For decoding we use a beam width of $10$ and a length penalty of $1.0$. 
We investigate the use of dropout \citep{JMLR:v15:srivastava14a} for the conditional baseline with rates from 10\% to 60\% in increments of 10\%. 
Best validation performance on WMT required a rate of 40\% for \textsc{En-De} and 50\% for \textsc{De-En}, while on IWSLT it required $50\%$ for either translation direction.
To spare resources, we also use these rates for training the simple \textsc{Joint} model. 

\begin{table}[t]
    \centering
    \begin{tabular}{l c c}
    \toprule
    & NC & IWSLT\\
    \midrule
    Dropout  & $30\%$ & $30\%$ \\
    Word dropout rate & $10\%$ & $20\%$ \\
    KL annealing steps & $80,000$ & $80,000$ \\ \midrule
    $\KL(q(z)||p(z))$ on \textsc{En-De} & $5.94$ & $8.01$ \\ \bottomrule
    \end{tabular}
    \caption{Strategies to promote use of latent representation along with the validation $\KL$ achieved.} 
    \label{tab:KL}
\end{table}

\paragraph{Avoiding collapsing to prior} Many have noticed that VAEs whose observation models are parameterised by \emph{strong generators}, such as recurrent neural networks, learn to ignore the latent representation \citep{bowman2016generating,higgins2016beta,LadderVAE,alemi18a}. In such cases, the approximate posterior ``collapses'' to the prior, and where one has a fixed prior, such as our standard Gaussian, this means that the posterior becomes independent of the data, which is obviously not desirable. \citet{bowman2016generating} proposed two techniques to counter this effect, namely, ``$\KL$ annealing'', and target word dropout. 
$\KL$ annealing consists in incorporating the $\KL$ term of Equation (\ref{eq:ELBO}) into the objective gradually, thus allowing the posterior to move away from the prior more freely at early stages of training. After a number of annealing steps, the $\KL$ term is incorporated in full and training continues with the actual ELBO. 
In our search we considered annealing for $20,000$ to $80,000$ training steps.  %
Word dropout consists in randomly masking words in observed target prefixes at a given rate. 
The idea is to harm the potential of the decoder to capitalise on correlations internal to the structure of the observation in the hope that it will rely more on the latent representation instead. We considered rates from $20\%$ to $40\%$ in increments of $10\%$. 
Table \ref{tab:KL} shows the configurations that achieve best validation results on \textsc{En-De}.
To spare resources, we reuse these hyperparameters for \textsc{De-En} experiments.
With these settings, we attain a non-negligible validation $\KL$ (see, last row of Table~\ref{tab:KL}), which indicates that the approximate posterior is different from the prior at the end of training. %

\paragraph{ELBO variants} We investigate the effect of conditioning on target observations for posterior inference during training against a simpler variant that conditions on the source alone. 
Table \ref{tab:r} suggests that conditioning on $\x$ is sufficient and thus we opt to continue with this simpler version.
Do note that when we use both observations for posterior inference, i.e. $q_\lambda(z|\x, \y)$, and thus train an approximation $r_\phi$ for prediction, we have additional parameters to estimate (e.g. due to the need to encode $\y$ for $q_\lambda$ and $\x$ for $r_\phi$), thus it may be the case that for these variants to show their potential we need larger data and/or prolonged training.

\begin{table}[t]
    \centering
    \begin{tabular}{l c}
    \toprule
    Objective & BLEU $\uparrow$ \\ \midrule
    $\ELBO_{\text{x,y}} - \KL(r_\phi(z|\x)||q_\lambda(z|\x, \y))$ & $14.7$\\ 
    $\ELBO_{\text{x,y}} - \KL(q_\lambda(z|\x, \y)||r_\phi(z|\x))$ & $14.8$\\ 
    $\ELBO_{\text{x,y}} - \JS(r_\phi(z|\x)||q_\lambda(z|\x, \y))$ & $14.9$\\
    $\ELBO_{\text{x,y}} - \left\Vert r_\phi(\x) - \mathbb E_{q_\lambda(z|\x, \y)}[Z] \right\Vert^2_2$ & $14.8$\\
    $\ELBO_{\text{x}}$  & $14.9$ \\
    \bottomrule
    \end{tabular}

    \caption{\textsc{En-De} validation results for NC training. $\ELBO_{\text{x}}$ means we condition on the source alone for posterior inference, i.e. the variational approximation $q_\lambda(z|\x)$ is used for training and for predictions. In all other cases, we condition on both observations for training, i.e. $q_\lambda(z|\x, \y)$, and train either a distribution $r_\phi(z|\x)$ or a point estimate $r_\phi(\x)$ for predictions.}
    \label{tab:r}
\end{table}

\subsection{\label{sec:results}Results}

\begin{table*}[t]
    \centering
    \begin{tabular}{l l c c c c} 
    \toprule
     & &  \multicolumn{2}{c}{\textsc{En-De}} & \multicolumn{2}{c}{\textsc{De-En}} \\
                \cmidrule(r{2pt}){3-4} \cmidrule(l{2pt}){5-6}
    Task & Model & BLEU $\uparrow$ & BEER $\uparrow$ & BLEU $\uparrow$ & BEER $\uparrow$ \\   \midrule
    IWSLT14 & \textsc{Cond}  & $23.0~(0.1)$ & $58.6~(0.1)$ & $27.3~(0.2)$ & $59.8~(0.1)$\\
    & \textsc{Joint}  &  $23.2$\phantom{$~(0.0)$} & $58.7$\phantom{$~(0.0)$} & $27.5$\phantom{$~(0.0)$} & $59.8$\phantom{$~(0.0)$}\\
    & \textsc{AEVNMT} & $\mathbf{23.4~(0.1)}$ & $\mathbf{58.8~(0.1)}$ & $\mathbf{28.0~(0.1)}$ &
    $\mathbf{60.1~(0.1)}$\\ 
    \midrule
    WMT16 & \textsc{Cond}  & $17.8~(0.2)$ & $53.1~(0.1)$ & $20.1~(0.1)$ & $\mathbf{53.7~(0.1)}$\\
    & \textsc{Joint}  & $17.9$\phantom{$~(0.0)$} & $53.4$\phantom{$~(0.0)$} & $20.1$\phantom{$~(0.0)$} & $\mathbf{53.7}$\phantom{$~(0.0)$}\\
    & \textsc{AEVNMT} & $\mathbf{18.4~(0.2)}$  &  $\mathbf{53.5~(0.1)}$ & $\mathbf{20.6~(0.2)}$ & $53.6~(0.1)$\\
    \bottomrule
    \end{tabular}
    \caption{\label{tab:in-domain}Test results for in-domain training on IWSLT (top) and NC (bottom): we report $\text{average}~(1\text{std})$ across $5$ independent runs for \textsc{Cond} and \textsc{AEVNMT}, but a single run of \textsc{Joint}.}
\end{table*}

In this section we report test results in terms of BLEU \citep{BLEU} and BEER \citep{BEER}, but in Appendix \ref{app:metrics} we additionally report METEOR \citep{METEOR} and TER \citep{TER}.
We de-truecase and de-tokenize our system's predictions and compute BLEU scores using SacreBLEU~\citep{Post18}.\footnote{Version string: \path{BLEU+case.mixed+numrefs.1+smooth.exp+tok.13a+version.1.2.12}} 
For BEER, METEOR and TER, we tokenize the results and test sets using the same tokenizer as used by SacreBLEU. We make use of BEER 2.0, and for METEOR and TER use \textsc{MultEval} \citep{multeval}.
In Appendix \ref{app:val} we report validation results, in this case in terms of BLEU alone as that is what we used for model selection. %
Finally, to give an indication of the degree to which results are sensitive to initial conditions (e.g. random initialisation of parameters), and to avoid possibly misleading signifiance testing, we report the average and standard deviation of $5$ independently trained models. %
To spare resources we do not report multiple runs for \textsc{Joint}, but our experience is that its performance varies similarly to that of the conditional baseline.

We start with the case where we can reasonably assume training data to be in-domain with respect to test data. 
Table \ref{tab:in-domain} shows in-domain training performance. 
First, we remark that our conditional baseline for the IWSLT14 task (IWSLT training) is very close to an external baseline trained on the same data \citep{bahdanau2016actor}.\footnote{\citet{bahdanau2016actor} report $27.56$ on the same test set for \textsc{De-En}, though note that they train on words rather than BPEs and use a different implementation of BLEU.}  %
The results on IWSLT show benefits from joint modelling and in particular from learning a shared latent space. For the WMT16 task (NC training), BLEU shows a similar trend, namely, joint modelling with a shared latent space (AEVNMT) outperforms both conditional modelling and the simple joint model. %

We now consider the scenario where we know for a fact that observations come from two different data distributions, which we realise by training our models on a concatenation of IWSLT and NC.
In this case, we perform model selection once on the concatenation of both development sets and evaluate the same model on each domain separately. 
We can see in Table \ref{tab:mix} %
that conditional modelling is never preferred,  \textsc{Joint} performs reasonably well, especially for \textsc{De-En}, and that in every comparison our AEVNMT outperforms the conditional baseline both in terms of BLEU and BEER.

\begin{table*}[t]
    \centering
    \begin{tabular}{l l c c c c}
    \toprule
     & & \multicolumn{2}{c}{WMT16} & \multicolumn{2}{c}{IWSLT14}\\
    \cmidrule(r{2pt}){3-4} \cmidrule(l{2pt}){5-6}
    Training & Model & BLEU $\uparrow$ & BEER $\uparrow$ & BLEU $\uparrow$ & BEER $\uparrow$  \\
    \midrule
    \textsc{En-De} & \textsc{Cond} & $17.6~(0.4)$ & $53.9~(0.2)$ & $23.9~(0.3)$ & $59.3~(0.1)$\\
    & \textsc{Joint} & $18.1$\phantom{$~(0.0)$} & $54.3$\phantom{$~(0.0)$} & $\mathbf{24.2}$\phantom{$~(0.0)$} & $\mathbf{59.5}$\phantom{$~(0.0)$}\\
    & \textsc{AEVNMT} & $\mathbf{18.4~(0.2)}$ & $\mathbf{54.5~(0.2)}$ & $24.1~(0.3)$ & $\mathbf{59.5~(0.2)}$\\
    \midrule
    \textsc{De-En} & \textsc{Cond} & $21.6~(0.2)$ & $55.5~(0.2)$ & $29.1~(0.2)$ & $60.9~(0.1)$\\
    & \textsc{Joint} & $\mathbf{22.3}$\phantom{$~(0.0)$} & $\mathbf{55.6}$\phantom{$~(0.0)$} & $\mathbf{29.2}$\phantom{$~(0.0)$} & $\mathbf{61.2}$\phantom{$~(0.0)$} \\
    & \textsc{AEVNMT} & $\mathbf{22.3~(0.1)}$ & $\mathbf{55.6~(0.1)}$ & $\mathbf{29.2~(0.1)}$ &  $61.1~(0.1)$\\
    \bottomrule
    \end{tabular}
    \caption{\label{tab:mix}Test results for mixed-domain training: we report $\text{average}~(1\text{std})$ across $5$ independent runs for \textsc{Cond} and \textsc{AEVNMT}, but a single run of \textsc{Joint}.} 
\end{table*}

Another common scenario where two very distinct data distributions are mixed is when we capitalise on the abundance of monolingual data and train on a concatenation of gold-standard bilingual data (we use NC) and synthetic bilingual data derived from target monolingual corpora via back-translation \citep{SennrichEtAl2016Mono} (we use News Crawl). In such a scenario the latent variable might be able to inform the translation model of the amount of noise present in the source sentence.
Table \ref{tab:NC+mono} shows results for both baselines and AEVNMT. First, note that synthetic data greatly improves the conditional baseline, in particular translating into English. 
Once again AEVNMT consistently outperforms conditional modelling and joint modelling without latent variables.

By mixing different sources of data we are trying to diagnose whether the generative model we propose is  robust to unknown and diverse sources of variation mixed together in one training set (e.g. NC + IWSLT or gold-standard + synthetic data). 
However, note that a point we are certainly not trying to make is that the model has been designed to perform \emph{domain adaptation}.
Nonetheless, in Appendix~\ref{apx:taus} we try to shed light on what happens when we use the model to translate genres it has never seen.
On a dataset covering various unseen genres, we observe that both \textsc{Cond} and \textsc{AEVNMT} perform considerably worse showing that without taking domain adaptation seriously both models are inadequate. In terms of BLEU, differences range from $-0.3$ to $0.8$ (\textsc{En-De}) and $0.3$ to $0.7$ (\textsc{De-En}) and are mostly in favour of AEVNMT (17/20 comparisons).
\paragraph{Remarks} It is intuitive to expect latent variable modelling to be most useful in settings containing high variability in the data, i.e. mixed-domain and synthetic data settings, though in our experiments \textsc{AEVNMT} shows larger improvements in the in-domain setting. We speculate two reasons for this: i) it is conceivable that variation in the mixed-domain and synthetic data settings are too large to be well accounted by a diagonal Gaussian; and ii) the benefits of latent variable modelling may diminish as the amount of available data grows.

\begin{table*}[t]
    \centering
    \begin{tabular}{l c c c c}
    \toprule
    WMT16 & \multicolumn{2}{c}{\textsc{En-De}} & \multicolumn{2}{c}{\textsc{De-En}}\\
    \cmidrule(r{2pt}){2-3} \cmidrule(l{2pt}){4-5}
    & BLEU $\uparrow$ & BEER $\uparrow$ & BLEU $\uparrow$ & BEER $\uparrow$  \\
    \midrule
    \textsc{Cond} & $17.8~(0.2)$ & $53.1~(0.1)$ & $20.1~(0.1)$ & $53.7~(0.1)$\\

    ~ + synthetic data &  $22.3~(0.3)$  & $\mathbf{57.0~(0.2)}$  & $26.9~(0.2)$  & $58.5~(0.1)$ \\
    
    \textsc{Joint} + synthetic data  & $22.2$\phantom{$~(0.0)$}  & $\mathbf{57.0}$\phantom{$~(0.0)$} & $26.7$\phantom{$~(0.0)$}  & $58.6$\phantom{$~(0.0)$} \\
    
    \textsc{AEVNMT} + synthetic data & $\mathbf{22.5~(0.2)}$ & $\mathbf{57.0~(0.1)}$  & $\mathbf{27.4~(0.2)}$  & $\mathbf{58.8~(0.1)}$ \\
    
    \bottomrule
    \end{tabular}
    \caption{\label{tab:NC+mono}Test results for training on NC plus  synthetic data (back-translated News Crawl): we report $\text{average}~(1\text{std})$ across $5$ independent runs for \textsc{Cond} and \textsc{AEVNMT}, but a single run of \textsc{Joint}.} %
    
\end{table*}

\subsection{Probing latent space}

To investigate what information the latent space encodes we explore the idea of training simple \emph{linear probes} or \emph{diagnostic classifiers} \citep{alain2016understanding,hupkes2018visualisation}. %
With simple Bayesian logistic regression we have managed to predict from $Z \sim q(z|\x)$ domain indicators (i.e. newswire vs transcripts) and gold-standard vs synthetic data at performance above $90\%$ accuracy on development set. 
However, a similar performance is achieved from the deterministic average state of the bidirectional encoder of the conditional baseline. 
We have also been able to predict from $Z \sim q(z|\x)$ the level of noise in back-translated data measured on the development set at the sentence level by an automatic metric, i.e. METEOR, with performance above what can be done with random features. 
Though again, the performance is not much better than what can be done with a conditional baseline. 
Still, it is worth highlighting that these aspects are rather coarse, and it is possible that the performance gains we report in \S\ref{sec:results} are due to far more nuanced variations in the data. 
At this point, however, we do not have a good qualitative assessment of this conjecture.

\section{Related Work}%

\paragraph{Joint modelling}

In similar work, \citet{shah2018generative} propose a joint generative model whose probabilistic formulation is essentially identical to ours. 
Besides some small differences in architecture, our work differs in two regards: motivation and strategy for predictions. 
Their goal is to jointly learn from multiple language pairs by sharing a single polyglot architecture \citep{johnson2016google}. %
Their strategy for prediction is based on a form of stochastic hill-climbing, where they sample an initial $z$ from the standard Gaussian prior and decode via beam search in order to obtain a draft translation $\tilde{\y} = \greedy_{\y} P(\y|z, \x)$.This translation is then iteratively refined by encoding the pair $\langle \x, \tilde \y \rangle$, re-sampling $z$, though this time from $q(z|\x, \tilde{\y})$, and re-decoding with beam search. 
Unlike our approach, this requires multiple calls to the inference network and to beam search. Moreover, the inference model, which is trained on gold-standard observations, is used on noisy target sentences.

\citet{cotterell2018explaining} interpret back-translation as a single iteration of a wake-sleep algorithm \citep{wakesleep} for a joint  model of bitext $P(\x, \y|\theta) = P(y|x, \theta) P_{\star}(\x)$.
They sample directly from the data distribution $P_\star(\x)$ and learn two NMT models, a generative $P(\y|\x, \theta)$ and an auxiliary model $Q(\x|\y, \phi)$, each trained on a separate objective. %
\citet{zhang2018joint} propose a joint model of bitext trained to incorporate the back-translation heuristic as a trainable component in a  formulation similar to that of \citet{cotterell2018explaining}.
In both cases, joint modelling is done without a shared latent space and without a source language model. %

\paragraph{Multi-task learning} An alternative to joint learning is to turn to multi-task learning and explore parameter sharing across models trained on different, though related, data with different objectives. 
For example, \citet{ChengEtAl2016SS} incorporate both source and target monolingual data by multi-tasking with a non-differentiable auto-encoding objective. They jointly train a source-to-target and target-to-source system that act as encoder and decoder respectively.
\citet{zhang-zong:2016:EMNLP2016} combine a source language model objective with a source-to-target conditional NMT objective and shared the source encoder in a multi-task learning fashion.

\paragraph{Variational LMs and NMT}
\citet{bowman2016generating} first proposed to augment a neural language model with a prior over latent space. Our source component is an instance of their model. More recently, \citet{SVAEText} proposed to use a hyperspherical uniform prior rather than a Gaussian and showed the former leads to better representations.
\citet{zhang-EtAl:2016:EMNLP20162} proposed the first VAE for NMT. They augment the conditional with a Gaussian sentence embedding %
and model observations as draws from the marginal $P(y|x, \theta) = \int p(z|x, \theta) P(y|x, z, \theta) \dd z$. 
Their formulation is a conditional deep generative model \citep{sohn2015learning} that does not model the source side of the data, where, rather than a fixed standard Gaussian, the latent model is itself parameterised and depends on the data. 
\citet{sdec} extend the model of  \citet{zhang-EtAl:2016:EMNLP20162} with a Markov chain of latent variables, one per timestep, allowing the model to capture greater variability.  %

\paragraph{Latent domains} In the context of statistical MT, \citet{cuong-simaan:2015:NAACL-HLT} estimate a joint distribution over sentence pairs while marginalising discrete latent domain indicators. 
Their model factorises over word alignments and is not used directly for translation, but rather to improve word and phrase alignments, or to perform data selection \citep{hoang-simaan:2014:Coling}, prior to training.  
There is a vast literature on domain adaptation for statistical machine translation \citep{SMTDomainSurvey}, as well as for NMT \citep{NMTDomainSurvey}, but a full characterisation of this exciting field is beyond the scope of this paper.

\section{Discussion and Future Work}

We have presented a joint generative model of translation data that generates both observations conditioned on a shared latent  representation.
Our formulation leads to questions such as \emph{why joint learning?} and \emph{why latent variable modelling?} to which we give an answer based on statistical facts about conditional modelling and marginalisation as well as empirical evidence of improved performance. Our model shows moderate but consistent improvements across various settings and over multiple independent runs. %

In future work, we shall investigate datasets annotated with demographics and personal traits in an attempt to assess how far we can go in capturing fine grained variation.
Though note that if such factors of variation vary widely in distribution, it may be na\"ive to expect we can model them well with a simple Gaussian prior. If that turns out to be the case, we will investigate mixing Gaussian components \citep{miao2016neural,srivastava2017autoencoding} and/or employing a hierarchical prior \citep{goyal2017NP}.

\section*{Acknowledgements}

\begin{wrapfigure}[4]{l}{0.26\linewidth}
\vspace{-12pt}
\includegraphics[width=0.15\textwidth]{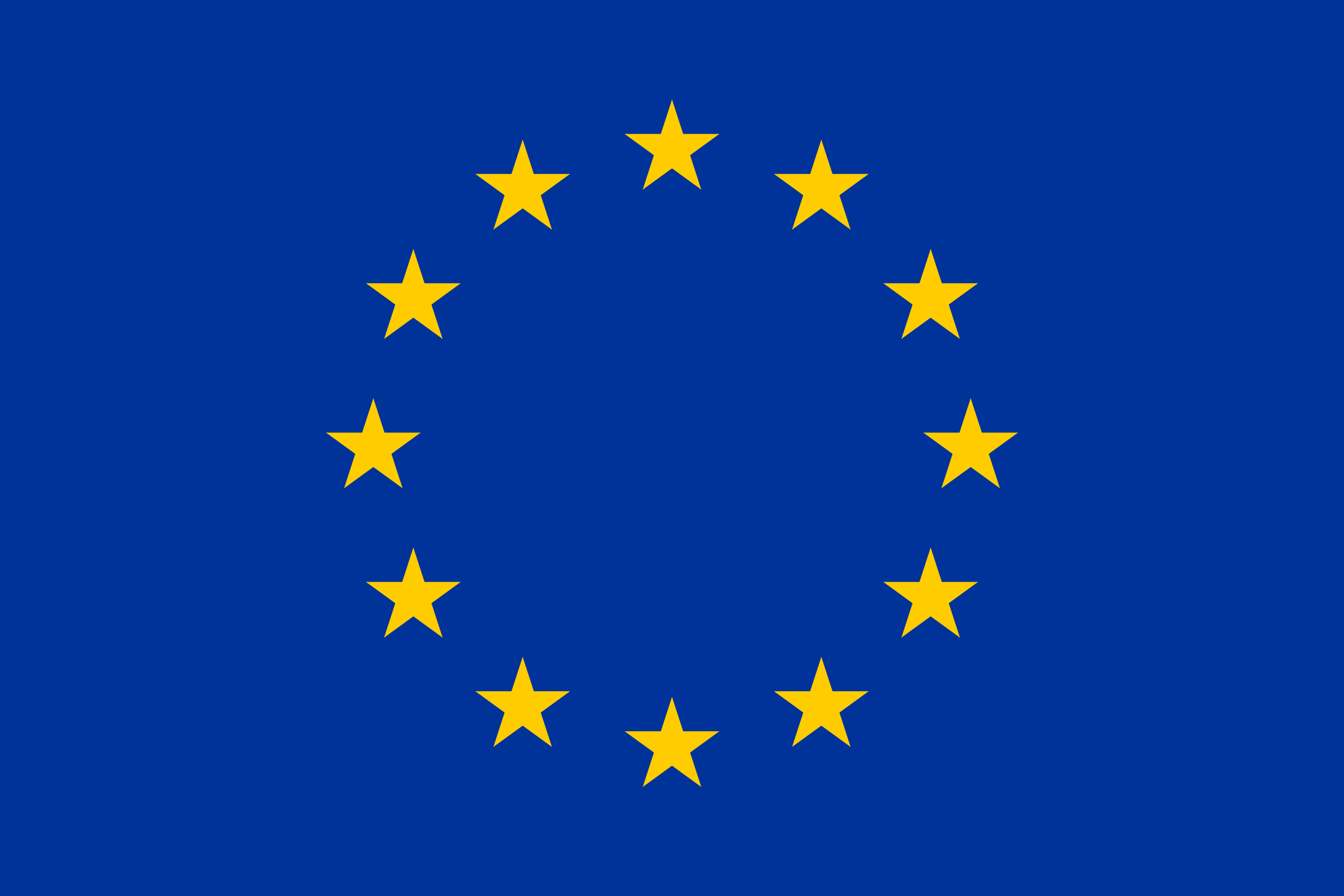}
\end{wrapfigure}
This project has received funding from the Dutch Organization for Scientific Research VICI Grant No 277-89-002 and from the European Union's Horizon 2020 research and innovation programme under grant agreement No 825299 (GoURMET). We also thank Philip Schulz, Khalil Sima'an, and Joost Bastings for comments and helpful discussions.
A Titan Xp card used for this research was donated by the NVIDIA Corporation.

\bibliography{jnmt}
\bibliographystyle{acl_natbib}

\clearpage
\newpage
\appendix

\section{\label{app:arch}Architectures}

Here we describe parameterisation of the different models presented in \S\ref{sec:model}. Rather than completely specifying standard blocks, we use the notation $\block(\text{inputs}; \text{parameters})$, where we give an indication of the relevant parameter set. This makes it easier to visually track  which model a component belongs to.

\subsection{Source Language Model}

The source language model consists of a sequence of categorical draws for $i=1, \ldots, \m$
\begin{align}
    X_i|z, \xh &\sim \Cat(g_\theta(z, \xh))
\end{align}
parameterised by a single-layer recurrent neural network using GRU units:
\begin{subequations}
\begin{align}
    \mathbf f_i &= \emb(x_i; \theta_{\text{emb-x}})  \label{eq:gen-emb-x}\\
    \mathbf h_0 &= \tanh(\affine(z; \theta_{\text{init-lm}})) \label{eq:init-lm}\\
    \mathbf h_i &= \GRU(\mathbf h_{i-1}, \mathbf f_{i-1}; \theta_{\text{gru-lm}}) \\
    g_\theta(z, \xh) &= \softmax(\affine(\mathbf h_i; \theta_{\text{out-x}})) ~.
\end{align}
\end{subequations}
We initialise the GRU cell with a transformation (\ref{eq:init-lm}) of the stochastic encoding $z$. %
For the simple joint model baseline we initialise the GRU with a vector of zeros as there is no stochastic encoding we can condition on in that case.

\subsection{Translation Model}
The translation model consists of a sequence of categorical draws for $j = 1, \ldots, \n$
\begin{equation}
    Y_j|z, \x, \yh \sim \Cat(f_\theta(z, \x, \yh))
\end{equation}
parameterised by an architecture that roughly follows \citet{BahdanauEtAl2015NMT}. 
The encoder is a bidirectional GRU encoder (\ref{eq:gen-bigru-x}) that shares source embeddings with the language model (\ref{eq:gen-emb-x}) and is initialised with its own projection of the latent representation put through a $\tanh$ activation. The decoder, also initialised with its own projection of the latent representation (\ref{eq:init-dec}), is a single-layer recurrent neural network with GRU units (\ref{eq:dec}). 
At any timestep the decoder is a function of the previous state, previous output word embedding, and a context vector. 
This context vector (\ref{eq:att}) is a weighted average of the bidirectional source encodings, of which the weights are computed by a Bahdanau-style attention mechanism. 
The output of the GRU decoder is projected to the target vocabulary size and mapped to the simplex using a $\softmax$ activation (\ref{eq:out-y}) to obtain the categorical parameters:
\begin{subequations}
\begin{align}
\mathbf s_0 &= \tanh(\affine(z; \theta_{\text{init-enc}}))\\
\mathbf s_1^m &= \BiGRU(\mathbf f_1^m, \mathbf s_0; \theta_{\text{bigru-x}})  \label{eq:gen-bigru-x}\\
\mathbf e_j &= \emb(y_j; \theta_{\text{emb-y}}) \\
\mathbf t_0 &= \tanh(\affine(z; \theta_{\text{init-dec}})) \label{eq:init-dec} \\
\mathbf c_j &= \attention(\mathbf s_1^m, \mathbf t_{j-1}; \theta_{\text{bahd}}) \label{eq:att} \\
\mathbf t_j &= \GRU(\mathbf t_{j-1}, [\mathbf c_j, \mathbf e_{j-1}]; \theta_{\text{gru-dec}}) ~ , \label{eq:dec} %
\end{align}
\end{subequations}
and 
\begin{equation}
    f_\theta(z, \x, \yh) = \softmax(\affine([\mathbf t_j, \mathbf e_{j-1}, \mathbf c_j]; \theta_{\text{out-y}})) ~. \label{eq:out-y}
\end{equation}
In baseline models, recurrent cells are initialised with a vector of zeros as there is no stochastic encoding we can condition on.

\subsection{\label{sec:infnet}Inference Network}

The inference model $q(z|\x, \y, \lambda)$ is a diagonal Gaussian
\begin{equation}
    Z|\x, \y \sim \mathcal N(\mathbf u, \diag(\mathbf s \odot \mathbf s))
\end{equation}
whose parameters are computed by an \emph{inference network}. 
We use two bidirectional GRU encoders to encode the source and target sentences separately. To spare memory, we reuse embeddings from the generative model (\ref{eq:emb-x}-\ref{eq:emb-y}), but we prevent updates to those parameters based on gradients of the inference network, which we indicate with the function $\detach$. 
To obtain fixed-size representations for the sentences, GRU encodings are averaged (\ref{eq:avg-x}-\ref{eq:avg-y}) .
\begin{subequations}
\begin{align}
\mathbf f_1^m &= \detach(\emb(x_1^m; \theta_{\text{emb-x}})) \label{eq:emb-x} \\
\mathbf e_1^n &= \detach(\emb(y_1^n; \theta_{\text{emb-y}})) \label{eq:emb-y}  \\
\mathbf h_x &= \avg\left(\BiGRU\left(\mathbf f_1^m; \lambda_{\text{gru-x}}\right)\right)  \label{eq:avg-x} \\
\mathbf h_y &= \avg\left(\BiGRU\left(\mathbf e_1^n; \lambda_{\text{gru-y}} \right)\right) \label{eq:avg-y}\\
\mathbf{h}_{xy} &= \concat(\mathbf h_x,  \mathbf h_y) \label{eq:concat} \\
\mathbf{h}_{u}  &= \ReLU(\affine(\mathbf h_{xy}; \lambda_{\text{u-hid}})) \\
\mathbf h_s &= \ReLU(\affine(\mathbf h_{xy}; \lambda_{\text{s-hid}}) \\
\mathbf u &= \affine(\mathbf{h}_{u}; \lambda_{\text{u-out}}) \label{eq:loc} \\
\mathbf s &= \softplus(\affine(\mathbf h_s; \lambda_{\text{s-out}}))  \label{eq:scale}
\end{align}
\end{subequations}
We use a concatenation $\mathbf h_{xy}$ of the average source and target encodings  (\ref{eq:concat}) as inputs to compute the parameters of the Gaussian approximate posterior, namely, $d$-dimensional location and scale vectors. Both transformations use ReLU hidden activations \citep{nair2010rectified}, but locations live in $\mathbb R^d$ and therefore call for linear output activations (\ref{eq:loc}), whereas scales live in $\mathbb R^d_{>0}$ and call for strictly positive outputs (\ref{eq:scale}), we follow \citet{ADVI} and use $\softplus$.
The complete set of parameters used for inference is thus $\lambda = \{\lambda_{\text{gru-x}}, \lambda_{\text{gru-y}}, \lambda_{\text{u-hid}}, \lambda_{\text{u-out}}, \lambda_{\text{s-hid}}, \lambda_{\text{s-out}}\}$.

\subsection{Prediction Network}

The \emph{prediction network} parameterises our prediction model $r(z|\x, \phi)$, a variant of the inference model that conditions on the source sentence alone. 
In \S\ref{sec:exp} we explore several variants of the ELBO using different parameterisations of $r_\phi$. 
In the simplest case we do not condition on the target sentence during training, thus we can use the same network both for training and prediction. The network is similar to the one described in \ref{sec:infnet}, except that there is a single bidirectional GRU and we use the average source encoding  (\ref{eq:avg-x}) as input to the predictors for $\mathbf u$ and $\mathbf s$ (\ref{eq:loc-x}-\ref{eq:scale-x}).
\begin{subequations}
\begin{align}
\mathbf h_u &= \ReLU(\affine(\mathbf h_{x}; \lambda_{\text{u-hid}})) \\
\mathbf h_s &= \ReLU(\affine(\mathbf h_{x}; \lambda_{\text{s-hid}})) \\
\mathbf u &= \affine(\mathbf h_u; \lambda_{\text{u-out}}) \label{eq:loc-x} \\
\mathbf s &= \softplus(\affine(\mathbf h_s; \lambda_{\text{s-out}}))  \label{eq:scale-x}
\end{align}
\end{subequations}

In all other cases we use $q(z|\x, \y, \lambda)$ parameterised as discussed in \ref{sec:infnet} for training, and design a separate network to parameterise $r_\phi$ for prediction. 
Much like the inference model, the prediction model is a diagonal Gaussian
\begin{equation}
    Z|\x \sim \mathcal N(\mathbf{\hat u}, \diag(\mathbf{\hat s} \odot \mathbf{\hat s}))
\end{equation}
also  parameterised by $d$-dimensional location and scale vectors, however in predicting $\mathbf{\hat u}$ and $\mathbf{\hat s}$ (\ref{eq:pred-loc}-\ref{eq:pred-scale}) it can only access an encoding of the source (\ref{eq:pred-avg-x}). 
\begin{subequations}
\begin{align}
\mathbf h_x &= \avg\left(\BiGRU\left(\mathbf f_1^m; \phi_{\text{gru-x}}\right)\right)  \label{eq:pred-avg-x} \\
\mathbf h_u &= \ReLU(\affine(\mathbf h_{x}; \phi_{\text{u-hid}})) \\
\mathbf h_s &= \ReLU(\affine(\mathbf h_{x}; \phi_{\text{s-hid}})) \\
\mathbf{\hat{u}}&= \affine(\mathbf h_u; \phi_{\text{u-out}}) \label{eq:pred-loc} \\
\mathbf{\hat s} &= \softplus(\affine(\mathbf h_s; \phi_{\text{s-out}}))  \label{eq:pred-scale}
\end{align}
\end{subequations}
The complete set of parameters is then $\phi = \{\phi_{\text{gru-x}}, \phi_{\text{u-hid}}, \phi_{\text{u-out}}, \phi_{\text{s-hid}}, \phi_{\text{s-out}}\}$.
For the deterministic variant, we use $\mathbf{\hat u}$ (\ref{eq:pred-loc}) alone to approximate $\mathbf u$ (\ref{eq:loc}), i.e. the posterior mean of $Z$.

\section{\label{app:stats}Graphical models}

\begin{figure}[h]
    \centering
    \begin{subfigure}{0.2\textwidth}
    \begin{tikzpicture}

    \node[obs]                        (y)     {$ y $};

    \node[obs, left = of y]           (x)     {$ x $};
    \node[above = of y]               (theta) {$ \theta $};
    \node[latent, below = of y]       (z)     {$ z $};

    \edge{x}{y};
    \edge{theta,z}{y,x};

    \plate {corpus} {(x) (y) (z)} {$ |\mathcal{D}| $};
    \end{tikzpicture}
    \caption{\label{fig:AEVNMT-gen}Generative model}
    \end{subfigure}
    ~
    \begin{subfigure}{0.2\textwidth}
    \begin{tikzpicture}

    \node[obs]                        (y)       {$ y $};
    \node[obs, left = of y]           (x)       {$ x $};
    \node[latent, below = of y]       (z)       {$ z $};
    \node[right = of z]               (lambda)     {$ \lambda $};
    \node[above = of y]               (phantom) {\phantom{$\theta$}};

    \edge{x,y}{z};
    \edge{lambda}{z};

    \plate {corpus} {(x) (y) (z)} {$ |\mathcal{D}| $};
    \end{tikzpicture}
    \caption{\label{fig:AEVNMT-inf}Inference model}
    \end{subfigure}
    \caption{On the left we have AEVNMT, a generative model parameterised by neural networks. On the right we show an independently parameterised model used for approximate posterior inference.} %
    \label{fig:AEVNMT}
\end{figure}
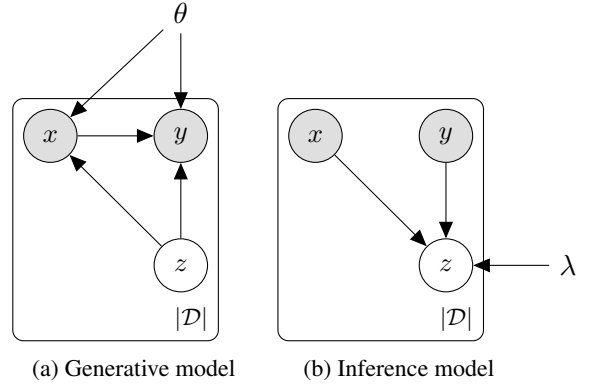

Figure \ref{fig:AEVNMT} is a graphical depiction of our AEVNMT model. Circled nodes denote random variables while uncircled nodes denote deterministic quantities. Shaded random variables correspond to observations and unshaded random variables are latent. The plate denotes a dataset of $|\mathcal D|$ observations.

\begin{figure}[h!]
    \centering
    \begin{subfigure}{\columnwidth}
    \centering
    \begin{tikzpicture}

    \node[obs]                        (x)      {$ x_i $};
    \node[obs, left = of x]           (xh)     {$ x_{<i} $};
    \node[obs, below = of x]                        (y)      {$ y_j $};
    \node[obs, below = of xh]           (yh)     {$ y_{<j} $};

    \node[latent, right = of x]       (z)     {$ z $};

    \edge{xh}{x};
    \edge{yh}{y};
    \edge{x}{y};
    \edge{z}{x,y};

    \plate {src} {(x) (xh)} {$ |x| \!$};
    \plate {tgt} {(y) (yh)} {$ |y| \!$};
    \end{tikzpicture}
    \caption{\label{fig:AEVNMT-joint}Joint distribution of AEVNMT}
    
    \vspace{10pt}
    
    \end{subfigure}

    \begin{subfigure}{\columnwidth}
    \centering
    \begin{tikzpicture}

    \node[obs]                        (y)      {$ y_j $};
    \node[obs, left = of y]           (yh)     {$ y_{<j} $};
    \node[obs, right = of y]           (yf)     {$ y_{>j} $};
    \node[obs, above = of y]       (x)      {$ x_i $};
    \node[obs, above = of yh]         (xh)      {$ x_{<i} $};
    \node[obs, above = of yf]         (xf)      {$ x_{>i} $};

    \edge[-]{y}{yh,yf,xh,xf};
    \edge[-]{x}{xh,xf,yh,yf};
    \edge[-]{x}{y};
    \plate {src} {(x) (xh) (xf)} {$ |x| $};
    \plate {sentence} {(y) (yh) (yf)} {$ |y| $};
    \end{tikzpicture}
    \caption{\label{fig:AEVNMT-marginal}Marginal distribution of AEVNMT}
    \end{subfigure}
    
    \vspace{10pt}
    
    \begin{subfigure}{\columnwidth}
    \centering
    \begin{tikzpicture}

    \node[obs]                        (x)      {$ x_i $};
    \node[obs, left = of x]           (xh)     {$ x_{<i} $};
    \node[obs, below = of x]                        (y)      {$ y_j $};
    \node[obs, below = of xh]           (yh)     {$ y_{<j} $};

    \edge{xh}{x};
    \edge{yh}{y};
    \edge{x}{y};

    \plate {src} {(x) (xh)} {$ |x| \!$};
    \plate {tgt} {(y) (yh)} {$ |y| \!$};
    \end{tikzpicture}
    \caption{\label{fig:JOINT}Joint distribution modelled without latent variables}
    \end{subfigure}

    \caption{Here we zoom in into the model of Figure \ref{fig:AEVNMT-gen} to show the statistical dependencies between observed variables. In the joint distribution (top), we have the directed dependency of a source word on all of the previous source words, and similarly, of a target word on all of the previous target words in addition to the complete source sentence. Besides, all observations depend directly on the latent variable $Z$. Marginalisation of $Z$ (middle) ties all variables together through undirected connections. At the bottom we show the distribution we get if we model the data distribution directly without latent variables.}
    \label{fig:AEVNMT-zoom}
\end{figure}
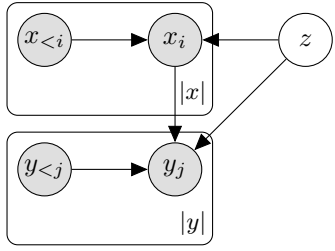
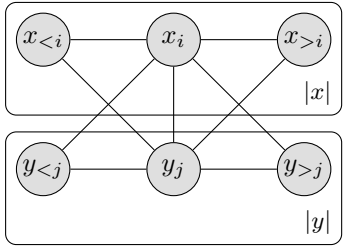
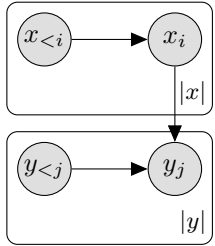

In Figure \ref{fig:AEVNMT-joint}, we illustrate the precise statistical assumptions of AEVNMT. Here plates iterate over words in either the source or the target sentence. Note that the arrow from $x_i$ to $y_j$ states that the $j$th target word depends on all of the source sentence, not on the $i$th source word alone, and that is the case because $x_i$ is within the source plate. 
In Figure \ref{fig:AEVNMT-marginal}, we illustrate the statistical dependencies induced in the marginal distribution upon marginalisation of latent variables. Recall that the marginal is the distribution which by assumption produced the observed data. Now compare that to the distribution modelled by the simple \textsc{Joint} model (Figure \ref{fig:JOINT}). 
Marginalisation induces undirected dependencies amongst random variables creating more structure in the marginal distribution. In graphical models literature this is known as \emph{moralisation} \citep{Koller+2009:PGM}.

\newpage

\newpage

\section{Robustness to out-of-domain data}
\label{apx:taus}

We use our stronger models, those trained on gold-standard NC bilingual data and synthetic News data, to translate test sets in various unseen genres. These data sets are collected and distributed by TAUS,\footnote{TAUS Hardware, TAUS Software, TAUS Industrial Electronics, TAUS Professional \& Business Services, and TAUS Legal available from TAUS data cloud \url{http://tausdata.org/}.} and have been used in scenarios of adapation to all domains at once \citep{HoangEtAl2016TACL}. 
Table~\ref{tab:TAUS} shows the performance of AEVNMT and the conditional baseline. The first thing to note is the remarkable drop in performance showing that without taking domain adaptation seriously both models are inadequate. In terms of BLEU, differences range from $-0.3$ to $0.8$ (\textsc{En-De}) and $0.3$ to $0.7$ (\textsc{De-En}) and are mostly in favour of AEVNMT, though note the increased standard deviations.

\begin{sidewaystable*}
    \centering
    \begin{tabular}{l l c c c c c c c c}
    \toprule
         & & \multicolumn{4}{c}{\textsc{En-De}} & \multicolumn{4}{c}{\textsc{De-En}}\\
         \cmidrule(r{2pt}){3-6} \cmidrule(r{2pt}){7-10}
        Model & Domain & BLEU $\uparrow$ & METEOR $\uparrow$ & TER $\downarrow$ & BEER $\uparrow$ & BLEU $\uparrow$ & METEOR $\uparrow$ & TER $\downarrow$ & BEER $\uparrow$  \\

    \midrule
        \textsc{Cond}
        & Computer Hardware & $8.7~(0.3)$ & $23.6~(0.2)$ & $82.3~(4.0)$ & $44.2~(0.3)$ & $13.0~(0.4)$ & $20.8~(0.1)$ & $73.5~(3.6)$ & $49.2~(0.1)$\\
        & Computer Software & $7.9~(0.2)$ & $22.9~(0.1)$ & $83.6~(3.2)$ & $44.1~(0.3)$ & $11.8~(0.8)$ & $20.4~(0.3)$ & $80.8~(7.0)$ & $\mathbf{48.6~(0.3)}$\\
        & Industrial Electronics & $7.6~(0.3)$ & $21.8~(0.2)$ & $89.6~(5.5)$ & $42.7~(0.3)$ & $10.1~(1.0)$ & $17.6~(0.4)$ & $85.7~(9.7)$ & $46.0~(0.4)$\\
        & Professional \& Business Services & $\mathbf{8.1~(0.4)}$ & $\mathbf{23.3~(0.4)}$ & $\mathbf{75.6~(0.4)}$ & $\mathbf{43.1~(0.4)}$ & $11.1~(0.2)$ & $20.5~(0.1)$ & $73.8~(1.4)$ & $44.9~(0.1)$\\
        & Legal & $12.3~(0.1)$ & $29.7~(0.3)$ & $75.3~(0.5)$ & $49.6~(0.1)$ & $14.2~(0.3)$ & $23.0~(0.3)$ & $67.3~(2.1)$ & $50.8~(0.3)$\\
    \midrule
        \textsc{AEVNMT}
        & Computer Hardware & $\mathbf{9.1~(0.3)}$ & $\mathbf{23.9~(0.4)}$ & $\mathbf{79.4~(1.3)}$ & $\mathbf{44.7~(0.2)}$ & $\mathbf{13.6~(0.3)}$ & $\mathbf{21.0~(0.2)}$ & $\mathbf{70.6~(1.4)}$ & $\mathbf{49.3~(0.2)}$\\
        & Computer Software & $\mathbf{8.3~(0.1)}$ & $\mathbf{23.2~(0.2)}$ & $\mathbf{81.4~(0.7)}$ & $\mathbf{44.5~(0.2)}$ & $\mathbf{12.2~(0.5)}$ & $\mathbf{20.7~(0.3)}$ & $\mathbf{77.0~(3.7)}$ & $\mathbf{48.6~(0.3)}$\\
        & Industrial Electronics & $\mathbf{8.1~(0.2)}$ & $\mathbf{22.2~(0.3)}$ & $\mathbf{85.2~(1.0)}$ & $\mathbf{43.1~(0.2)}$ & $\mathbf{10.8~(0.6)}$ & $\mathbf{17.9~(0.3)}$ & $\mathbf{79.9~(4.9)}$ & $\mathbf{46.3~(0.2)}$\\
        & Professional \& Business Services & $7.8~(0.7)$ & $23.0~(0.7)$ & $75.7~(1.0)$ & $43.0~(0.5)$ & $\mathbf{11.5~(0.2)}$ & $\mathbf{20.6~(0.2)}$ & $\mathbf{73.1~(0.7)}$ & $\mathbf{45.1~(0.1)}$\\
        & Legal & $\mathbf{13.1~(0.2)}$ & $\mathbf{30.5~(0.3)}$ & $\mathbf{74.4~(0.6)}$ & $\mathbf{49.9~(0.2)}$ & $\mathbf{14.5~(0.2)}$ & $\mathbf{23.2~(0.2)}$ & $\mathbf{66.1~(1.3)}$ & $\mathbf{50.9~(0.3)}$\\

    \bottomrule
    \end{tabular}
    \caption{\label{tab:TAUS}Performance of models trained with NC and back-translated News on various TAUS test sets: we report $\text{average}~(1\text{std})$ across $5$ independent runs.}
\end{sidewaystable*}

\newpage %
\onecolumn{\section{\label{app:val}Validation results}}
    \vspace{40pt}

\begin{table*}[h]
    \centering
    \begin{tabular}{l c c c c} 
    \toprule
    & \multicolumn{2}{c}{WMT16} & \multicolumn{2}{c}{IWSLT14}\\
    \cmidrule(r{2pt}){2-3} \cmidrule(l{2pt}){4-5}
    & \textsc{En-De} & \textsc{De-En} & \textsc{En-De} & \textsc{De-En} \\
    \midrule
    \textsc{Cond} & $14.5~(0.2)$ & $16.9~(0.2)$ & $25.1~(0.1)$ & $30.8~(0.1)$  \\
    \textsc{Joint} & $\mathbf{14.8}$\phantom{$~(0.0)$} & $17.1$\phantom{$~(0.0)$} & $25.2$\phantom{$~(0.0)$} & $31.0$\phantom{$~(0.0)$} \\
    \textsc{AEVNMT} & $\mathbf{14.8~(0.2)}$ & $\mathbf{17.4~(0.2)}$ & $\mathbf{25.7~(0.0)}$ & $\mathbf{31.4~(0.0)}$ \\ 
    \bottomrule
    \end{tabular}
    \caption{\label{tab:in-domain-val}Validation results reported in BLEU for in-domain training on NC and IWSLT: we report $\text{average}~(1\text{std})$ across $5$ independent runs for \textsc{Cond} and \textsc{AEVNMT}, but a single run of \textsc{Joint}.} 

    \vspace{40pt}
    
    \begin{tabular}{l c c}
    \toprule
    WMT \& IWSLT & \textsc{En-De} & \textsc{De-En}\\
    \midrule
    \textsc{Cond} & $20.5~(0.1)$ & $25.9~(0.1)$\\
    \textsc{Joint} & $20.7$\phantom{$~(0.0)$} & $\mathbf{26.1}$\phantom{$~(0.0)$}\\
    \textsc{AEVNMT} & $\mathbf{20.8~(0.1)}$ & $\mathbf{26.1~(0.1)}$\\
    \bottomrule
    \end{tabular}
    \caption{\label{tab:mixed-domain=-val}Validation results reported in BLEU for mixed-domain training: we report $\text{average}~(1\text{std})$ across $5$ independent runs for \textsc{Cond} and \textsc{AEVNMT}, but a single run of \textsc{Joint}. The validation set used is a concatenation of the development sets from WMT and IWSLT.}
    
    \vspace{40pt}

    \begin{tabular}{l c c}
    \toprule
    WMT16 & \textsc{En-De} & \textsc{De-En}\\
    \midrule
    \textsc{Cond} & $14.5~(0.2)$ & $16.9~(0.2)$\\
    ~ + synthetic data & $17.4~(0.1)$ & $21.8~(0.1)$\\
    \textsc{Joint} + synthetic data & $17.3$\phantom{$~(0.0)$} & $21.8$\phantom{$~(0.0)$}\\
    \textsc{AEVNMT} + synthetic data & $\mathbf{17.6~(0.1)}$ & $\mathbf{22.1~(0.1)}$\\
    \bottomrule
    \end{tabular}
    \caption{\label{tab:back-translation-val}Validation results reported in BLEU for training on NC plus synthetic data: we report $\text{average}~(1\text{std})$ across $5$ independent runs for \textsc{Cond} and \textsc{AEVNMT}, but a single run of \textsc{Joint}.}
\end{table*}

\clearpage
\section{\label{app:metrics}Additional Metrics}

\begin{sideways}%
\begin{minipage}{1.4\textwidth} %
    \vspace{48pt} %
    
    \hspace{30pt} %
    \begin{tabular}{l c c c c c c c c} 
    \toprule
    IWSLT14 &       \multicolumn{4}{c}{\textsc{En-De}} & \multicolumn{4}{c}{\textsc{De-En}} \\
                \cmidrule(r{2pt}){2-5} \cmidrule(l{2pt}){6-9}
                 & BLEU $\uparrow$ &  METEOR $\uparrow$ & TER $\downarrow$ & BEER $\uparrow$ & BLEU $\uparrow$ & METEOR $\uparrow$ & TER $\downarrow$ & BEER $\uparrow$  \\   \midrule
    \textsc{Cond}  & $23.0~(0.1)$ & $42.4~(0.1)$ & $56.0~(0.1)$ & $58.6~(0.1)$ & $27.3~(0.2)$ & $30.3~(0.1)$ & $52.4~(0.5)$ & $59.8~(0.1)$\\
    \textsc{Joint}  &  $23.2$\phantom{$~(0.0)$} & $\mathbf{42.8}$\phantom{$~(0.0)$} & $56.1$\phantom{$~(0.0)$} & $58.7$\phantom{$~(0.0)$} & $27.5$\phantom{$~(0.0)$} & $30.3$\phantom{$~(0.0)$} & $52.7$\phantom{$~(0.0)$} & $59.8$\phantom{$~(0.0)$}\\
    \textsc{AEVNMT} & $\mathbf{23.4~(0.1)}$ & $\mathbf{42.8~(0.2)}$ & $\mathbf{55.5~(0.3)}$ & $\mathbf{58.8~(0.1)}$ & $\mathbf{28.0~(0.1)}$ &
    $\mathbf{30.6~(0.1)}$ & $\mathbf{51.2~(0.6)}$ & $\mathbf{60.1~(0.1)}$\\ 
    \bottomrule
    \end{tabular}
    \captionof{table}{\label{tab:app-IWSLT}Test results for in-domain training on IWSLT: we report $\text{average}~(1\text{std})$ across $5$ independent runs for \textsc{Cond} and \textsc{AEVNMT}, but a single run of \textsc{Joint}.}
    
    \vspace{20pt}
    
    \hspace{30pt} %
    \begin{tabular}{l c c c c c c c c} 
    \toprule
    WMT16 &       \multicolumn{4}{c}{\textsc{En-De}} & \multicolumn{4}{c}{\textsc{De-En}} \\
                \cmidrule(r{2pt}){2-5} \cmidrule(l{2pt}){6-9}
                 & BLEU $\uparrow$ &  METEOR $\uparrow$ & TER $\downarrow$ & BEER $\uparrow$ & BLEU $\uparrow$ & METEOR $\uparrow$ & TER $\downarrow$ & BEER $\uparrow$ \\   \midrule
    \textsc{Cond}  & $17.8~(0.2)$ & $35.9~(0.2)$ & $65.2~(0.4)$ & $53.1~(0.1)$ & $20.1~(0.1)$ &  $\mathbf{26.0~(0.1)}$ & $62.0~(0.3)$ & $\mathbf{53.7~(0.1)}$\\
    \textsc{Joint}  & $17.9$\phantom{$~(0.0)$} & $36.2$\phantom{$~(0.0)$} & $64.1$\phantom{$~(0.0)$} & $53.4$\phantom{$~(0.0)$} & $20.1$\phantom{$~(0.0)$} & $\mathbf{26.0}$\phantom{$~(0.0)$} & $62.7$\phantom{$~(0.0)$} & $\mathbf{53.7}$\phantom{$~(0.0)$}\\
    \textsc{AEVNMT} & $\mathbf{18.4~(0.2)}$  & $\mathbf{36.6~(0.2)}$ & $\mathbf{64.0~(0.4)}$ & $\mathbf{53.5~(0.1)}$ & $\mathbf{20.6~(0.2)}$ & $\mathbf{26.0~(0.1)}$ & $\mathbf{60.7~(0.8)}$ & $53.6~(0.1)$\\
    \bottomrule
    \end{tabular}
    \captionof{table}{\label{tab:app-NC}Test results for in-domain training on NC: we report $\text{average}~(1\text{std})$ across $5$ independent runs for \textsc{Cond} and \textsc{AEVNMT}, but a single run of \textsc{Joint}.} 
    
    \vspace{20pt}
    
    \begin{tabular}{l c c c c c c c c}
    \toprule
    WMT16 & \multicolumn{4}{c}{\textsc{En-De}} & \multicolumn{4}{c}{\textsc{De-En}}\\
    \cmidrule(r{2pt}){2-5} \cmidrule(l{2pt}){6-9}
    & BLEU $\uparrow$ & METEOR $\uparrow$ & TER $\downarrow$ & BEER $\uparrow$ & BLEU $\uparrow$ & METEOR $\uparrow$ & TER $\downarrow$ & BEER $\uparrow$  \\
    \midrule
    
    \textsc{Cond} & $17.8~(0.2)$ & $35.9~(0.2)$ & $65.2~(0.4)$ & $53.1~(0.1)$ & $20.1~(0.1)$ & $26.0~(0.1)$ & $62.0~(0.3)$ & $53.7~(0.1)$\\

    ~ + synthetic data &  $22.3~(0.3)$  & $40.9~(0.2)$ & $58.5~(0.5)$ & $\mathbf{57.0~(0.2)}$ & $26.9~(0.2)$  & $30.4~(0.1)$ & $53.0~(0.5)$ & $58.5~(0.1)$ \\
    
    \textsc{Joint} + synthetic data  & $22.2$\phantom{$~(0.0)$}  & $40.8$\phantom{$~(0.0)$} & $\mathbf{58.1}$\phantom{$~(0.0)$} & $\mathbf{57.0}$\phantom{$~(0.0)$} & $26.7$\phantom{$~(0.0)$}  & $30.2$\phantom{$~(0.0)$} & $52.1$\phantom{$~(0.0)$} & $58.6$\phantom{$~(0.0)$} \\
    
    \textsc{AEVNMT} + synthetic data & $\mathbf{22.5~(0.2)}$ & $\mathbf{41.0~(0.1)}$ & $\mathbf{58.1~(0.2)}$ & $\mathbf{57.0~(0.1)}$ & $\mathbf{27.4~(0.2)}$ & $\mathbf{30.6~(0.1)}$ & $\mathbf{52.0~(0.1)}$ & $\mathbf{58.8~(0.1)}$ \\
    
    \bottomrule
    \end{tabular}
    \captionof{table}{\label{tab:app-NC+mono}Test results for training on NC plus  synthetic data (back-translated News Crawl): we report $\text{average}~(1\text{std})$ across $5$ independent runs for \textsc{Cond} and \textsc{AEVNMT}, but a single run of \textsc{Joint}.}

\end{minipage}
\end{sideways}

\begin{sidewaystable*}
    \centering
    
    \begin{tabular}{l l c c c c c c c c}
    \toprule
    \textsc{En-De} & & \multicolumn{4}{c}{WMT16} & \multicolumn{4}{c}{IWSLT14}\\
    \cmidrule(r{2pt}){3-6} \cmidrule(l{2pt}){7-10}
    Training & Model & BLEU $\uparrow$ & METEOR $\uparrow$ & TER $\downarrow$ & BEER $\uparrow$ & BLEU $\uparrow$ & METEOR $\uparrow$ & TER $\downarrow$ & BEER $\uparrow$  \\
    \midrule
    \textsc{En-De} & \textsc{Cond} & $17.6~(0.4)$ & $35.7~(0.3)$ & $61.9~(0.7)$ & $53.9~(0.2)$ & $23.9~(0.3)$ & $43.1~(0.3)$ & $54.7~(0.2)$ & $59.3~(0.1)$\\
    
    & \textsc{Joint} & $18.1$\phantom{$~(0.0)$} & $36.3$\phantom{$~(0.0)$} & $61.4$\phantom{$~(0.0)$} & $54.3$\phantom{$~(0.0)$} & $\mathbf{24.2}$\phantom{$~(0.0)$} & $43.4$\phantom{$~(0.0)$} & $54.5$\phantom{$~(0.0)$} & $\mathbf{59.5}$\phantom{$~(0.0)$}\\

    & \textsc{AEVNMT} & $\mathbf{18.4~(0.2)}$ & $\mathbf{36.6~(0.3)}$ & $\mathbf{60.9~(0.4)}$ & $\mathbf{54.5~(0.2)}$ & $24.1~(0.3)$ & $\mathbf{43.5~(0.3)}$ & $\mathbf{54.0~(0.3)}$ & $\mathbf{59.5~(0.2)}$\\
    
    \midrule 
    
    \textsc{De-En} & \textsc{Cond} & $21.6~(0.2)$ & $27.4~(0.1)$ & $59.9~(0.7)$ & $55.5~(0.2)$ & $29.1~(0.2)$ & $\mathbf{31.5~(0.1)}$ & $50.9~(0.5)$ & $60.9~(0.1)$\\
    
    & \textsc{Joint} & $\mathbf{22.3}$\phantom{$~(0.0)$} & $27.4$\phantom{$~(0.0)$} & $58.8$\phantom{$~(0.0)$} & $\mathbf{55.6}$\phantom{$~(0.0)$} & $\mathbf{29.2}$\phantom{$~(0.0)$} & $\mathbf{31.5}$\phantom{$~(0.0)$} & $\mathbf{49.2}$\phantom{$~(0.0)$} & $\mathbf{61.2}$\phantom{$~(0.0)$}\\
    
    & \textsc{AEVNMT} & $\mathbf{22.3~(0.1)}$ & $\mathbf{27.5~(0.1)}$ & $\mathbf{57.8~(0.6)}$ & $\mathbf{55.6~(0.1)}$ & $\mathbf{29.2~(0.1)}$ & $\mathbf{31.5~(0.0)}$ & $\mathbf{49.2~(0.4)}$ & $61.1~(0.1)$\\

    \bottomrule
    \end{tabular}
    \caption{\label{tab:app-mix}Test results for mixed-domain training: we report $\text{average}~(1\text{std})$ across $5$ independent runs for \textsc{Cond} and \textsc{AEVNMT}, but a single run of \textsc{Joint}.}

\end{sidewaystable*}

\end{document}